\documentclass[10pt,twocolumn,letterpaper]{article}

\usepackage[pagenumbers]{cvpr} 

\definecolor{cvprblue}{rgb}{0.21,0.49,0.74}
\usepackage[pagebackref,breaklinks,colorlinks,allcolors=cvprblue]{hyperref}
\usepackage{cuted}     
\usepackage{caption}   
\def\paperID{*****} 
\def\confName{CVPR}
\def\confYear{2027}

\title{When Does Test-Time Search Help World Action Models? An Empirical Audit of Candidate Selection and Simulator Integrity}

\author{
  \textbf{Jun Liu\textsuperscript{1}},
  \textbf{Dajie Sun\textsuperscript{2}},
   \textbf{Zhenglun Kong\textsuperscript{3}},
   \textbf{Zhenjiang Fan\textsuperscript{4}},
  \textbf{Peiyan Dong\textsuperscript{5}},
  \textbf{Fan Yang\textsuperscript{6}},\\
  \textbf{Wenbin Zhang \textsuperscript{7}},
  \textbf{Geng Yuan\textsuperscript{8}},
  \textbf{Xue Lin\textsuperscript{9}},
  \textbf{Yanzhi Wang\textsuperscript{9,10},
  \textbf{Dong Huang\textsuperscript{1}}
  }
\\
\\
  \textsuperscript{1}CMU,
  \textsuperscript{2}UC Berkeley, 
  \textsuperscript{3}Harvard University,
  \textsuperscript{4}Stanford  University,
  \textsuperscript{5}MIT ,
\\
 \textsuperscript{6}Fujitsu, 
  \textsuperscript{7}Florida International University,
  \textsuperscript{8}University of Georgia,
\\
  \textsuperscript{9}Northeastern University 
  \textsuperscript{10}EmbodyX Inc
 \\ 
}

\begin{document}
\maketitle
\maketitle

\begin{strip}
  \centering
  \includegraphics[width=0.90\linewidth]{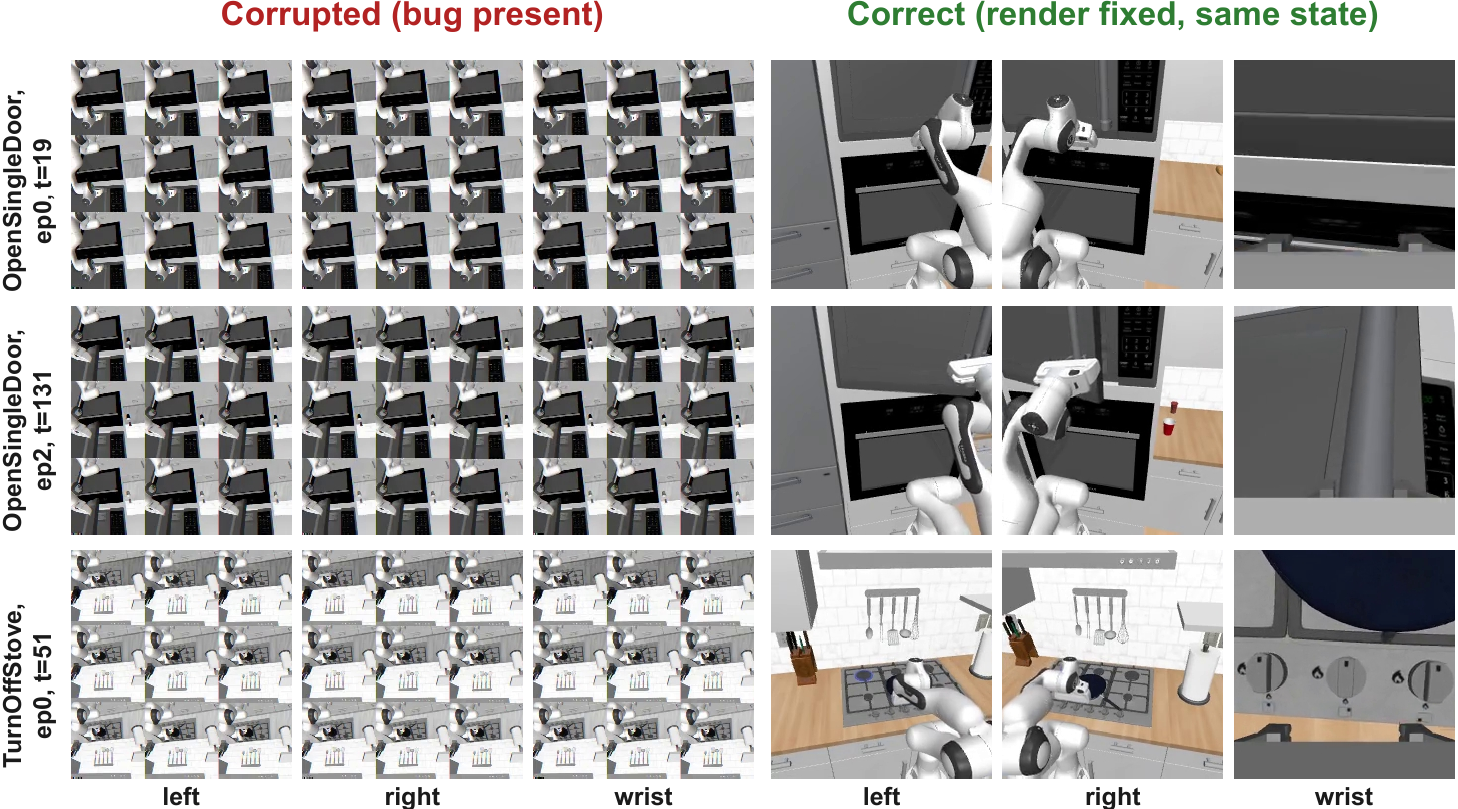}
  \captionsetup{type=figure,hypcap=false}
  \captionof{figure}{
  The silent rendering corruption, captured on the planning
inference path, is reproducible across episodes and tasks. Each row is
one example (task, episode, time step); the left column is the corrupted
frame, a $3\times3$ tiled mosaic of the framebuffer, and the right
column is the render-fixed rerun of the same configuration at the same
step, which agrees with the bug run to within encoding noise before
corruption onset and therefore depicts the same physical state. No error
is raised when the corruption occurs. Frames are the full $224\times672$
observation (left, right, and wrist cameras). These frames are from the
planning path, where corruption is frequent; the official direct-policy
benchmark, where it is rare, is analyzed separately in Section~\ref{sec:reliability}.
  }
  \label{fig:corruption}
\end{strip}

\begin{abstract}
World-action models increasingly apply test-time scaling, spending extra
inference computation on candidate selection and adaptive allocation to
improve control. Such gains are credible only when candidates carry
exploitable signal, nominal compute savings become real speedup, and
evaluation stays reliable. We audit all three on Cosmos Policy and
RoboCasa with a premise-first, pre-registered methodology. On clean data,
a learned value selector is 1.66 pp worse than random; a hindsight oracle
gains only 1.40 pp, with an episode-level interval spanning zero; and
82\% of states show identical candidate outcomes, leaving little
selection headroom. Token pruning cuts per-forward FLOPs by up to 47\% but yields at most a measured 1.15$\times$ end-to-end speedup on the deployed path (1.25$\times$ on a
decode-excluded path), short of a 1.5$\times$ target, and our
adaptive-allocation experiment is inconclusive under a power gate. We
further identify and fix a silent rendering corruption whose rate rises
sharply along compute-heavy planning paths while leaving the official
direct-policy result unaffected, and release a detector, fix, and
scripts. These findings show why test-time scaling should be audited as a
chain of premises (decision signal, realized efficiency, evaluation
integrity) rather than by success rate alone. Conclusions are scoped to
the tested configuration.
\end{abstract}    
\section{Introduction}

\label{sec:intro}

\begin{figure*}[tb]
\centering
\includegraphics[width=0.9\linewidth]{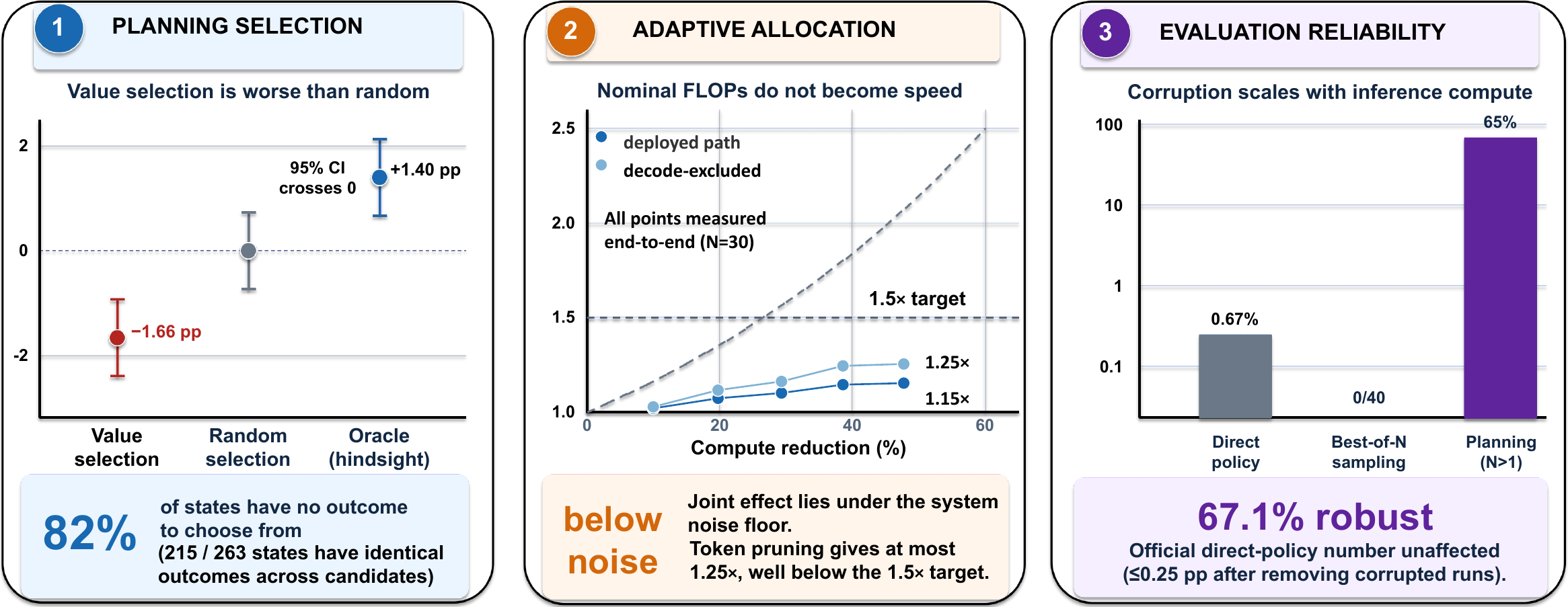}
\caption{A premise-first audit of test-time scaling in Cosmos Policy
on RoboCasa. \textbf{(1) Planning selection.} Limited variation in
candidate outcomes constrains the headroom for selection, and the
deployable value selector underperforms random selection, while the
hindsight oracle provides only modest evidence of improvement.
\textbf{(2) Adaptive allocation.} Substantial nominal FLOPs reductions
translate into modest end-to-end latency gains, while the joint
allocation effect remains unresolved at the available noise and
statistical power.
\textbf{(3) Evaluation reliability.} Rendering corruption becomes much
more frequent in the compute-intensive planning path than under
direct-policy or best-of-$N$ evaluation, while the official direct-policy
result remains essentially unaffected. All findings are scoped to the
tested Cosmos Policy--RoboCasa configuration.}
\label{fig:teaser}
\vspace{-5pt}
\end{figure*}

World-action models (WAMs) such as Cosmos Policy~\cite{kim2026cosmos}
adapt large pretrained video models into robot policies that jointly
generate actions, future states, and values within a shared latent
diffusion process. A central promise of this design is \emph{test-time
scaling}: by generating candidate actions, predicting their future
consequences, and evaluating them with a value function, WAMs can
allocate additional inference-time computation to identify promising
actions. Related peer-reviewed work has explored complementary aspects
of this problem, including model-based planning with action-conditioned
video world models~\cite{zhu2025irasim} and efficient diffusion-based
robot policies through one-step distillation~\cite{3780338.3782852},
training-free adaptive caching~\cite{ji2026bac}, and rollout-adaptive
pruning and activation reuse~\cite{ji2026sag}.
 
Yet whether extra inference-time computation improves control remains
insufficiently examined. More computation helps only if it exposes
decision-relevant candidate differences, produces practical end-to-end
effects, and is evaluated reliably; success rates alone cannot distinguish
a weak selection signal, unrealized compute savings, and simulator artifacts
~\cite{li2024evaluating}. We therefore audit test-time scaling as a chain
of premises rather than a single performance number: is useful signal
present, do nominal savings translate into end-to-end gains, and does the
evaluation path remain reliable under added computation?

We instantiate this audit on Cosmos Policy and RoboCasa~\cite{robocasa365}.
Rather than proposing a planner, we combine paired candidate comparisons
and a deployment-conditioned hindsight oracle, pre-registered decision
criteria and a power gate, complete-query latency measurements, and a
framebuffer-sentinel integrity check.
 
The audit yields three findings.
 
\noindent\textbf{(1) Candidate selection is limited by the available
signal, not just by the selector.}
On render-verified clean data, the deployable value selector performs
$1.66$\,pp worse than random, and even a hindsight oracle (an upper bound)
gains only $1.40$\,pp with an interval including zero, because in $82\%$ of
query states all candidates share an outcome. This does not show selection
is generally ineffective; it shows selector quality must be read alongside
the separability of the candidates it receives.
 
\noindent\textbf{(2) Nominal compute reduction is not a reliable proxy
for practical acceleration.}
Directly timing the complete query, token pruning yields at most a
measured $1.15\times$ end-to-end speedup on the deployed path
($1.25\times$ with future-image decode excluded), below the pre-specified
$1.5\times$ target, because the gain is bounded by the irreducible
autoencoder encode and decode. A separate joint-allocation diagnostic
falls below the measurement's noise resolution and did not clear its power
gate, so it is inconclusive rather than evidence of no effect.
 
\noindent\textbf{(3) Evaluation integrity can depend on the inference
path being evaluated.}
We identify and fix a silent rendering corruption in the
RoboCasa/\texttt{robosuite} stack whose frequency is small under the
official direct-policy protocol but rises sharply in the compute-intensive
planning path. It is more than a generic simulator bug: changing
inference-time computation changes exposure to a measurement failure, so
the compute-intensive paths that realize test-time scaling are the most
exposed.
 
Together, these findings make the contribution a testable audit
methodology and a configuration-specific case study, not a claim that
test-time scaling fails for all WAMs.

\section{Related Work}
\label{sec:related}

\noindent\textbf{World models for control.}
World models that predict action-conditioned futures have advanced
rapidly, spanning diffusion-based video prediction
\cite{alonso2024diffusion, guo2024prediction,chun2026sparse}, 3D and persistent
representations \cite{zhou2026learning, lu2025gwm}, and
manipulation-specific models \cite{kim2026dexterous}. The backbone we
study, Cosmos Policy~\cite{kim2026cosmos}, follows the joint action-video line in which
actions, futures, and values are produced within a single denoising
process \cite{guo2024prediction}. We do not propose a new
world model; we audit whether the inference-time computation layered on
top of such a model improves control.

\noindent\textbf{Efficient world-model inference.}
Because prediction and planning in these models are costly, a growing
body of work reduces their inference cost, through sparse imagination of
rollouts \cite{chun2026sparse}, disentangled dynamics prediction
\cite{yin2026ddpwm}, efficient on-policy rollout in diffusion world
models \cite{cohen2026horizon}, and warm-started prediction that reuses
spatiotemporal consistency \cite{li2026step}. These methods report
nominal compute reductions. We audit whether such reductions convert to
end-to-end latency on a representative stack, and find that token-level
savings are capped by an irreducible encoding fraction rather than
realized as speedup, and that the joint allocation effect lies below the
measurement noise floor. X-WAM~\cite{guo2026xwam} audit underscores why architectural avenues can be more fruitful than token pruning alone for practical speedups.

\noindent\textbf{Test-time scaling and candidate selection.}
A parallel line scales inference by sampling and selecting candidate
actions, through verifier-free scoring \cite{jang2026verifier}, relative
action critics \cite{li2026vlaattc}, and reward-ranked world-model
rollouts \cite{zhang2026worldinworld}; analogous strategies in text and
digital-agent settings use self-verification, process reward models, and
execution-free critics \cite{zhao2025sample, bilal2026what,
hua2026learning, shum2026swe,jang2026verifier}. These works propose selectors and report
positive scaling. We instead audit whether the selection signal exists,
and find a learned value selector worse than random on a strong base
policy where candidate outcomes rarely differ. 

\noindent\textbf{Reliability of policy evaluation.}
Recent work asks whether simulator- and world-model-based evaluation is
trustworthy, training learned evaluators to match real rollouts
\cite{li2026scaling}, testing robustness under real-to-sim translation
\cite{jangir2026robotarena}, and reducing evaluation cost while
preserving rankings \cite{chen2026manipevalagent}. These address the
fidelity of a learned evaluator or a synthesized scene. Our reliability
finding is at the level of the evaluation stack itself: a silent
rendering corruption in the standard RoboCasa and \texttt{robosuite}
pipeline whose rate rises with per-query inference compute, for which we
release a detector and a fix.

\section{A Diagnostic Methodology for Test-Time Compute}
\label{sec:methodology}
Because extra inference compute can change outputs without improving
control, success-rate movements alone are not evidence of scaling gains.
Our pre-registered, falsification-first audit tests for usable candidate
signal and reliable evaluation before interpreting success
(Section~\ref{sec:reliability}).

\subsection{Verify the Premise Before Measuring the Gain}
\label{subsec:premise-first}

Before evaluating an inference-time mechanism, we test whether its
required signal exists. For best-of-$N$ planning, candidate outcomes must
differ in ways a selector can recover; we test this directly in
Section~\ref{sec:planning-audit}, rather than infer it from downstream success.
Supplement Figure~\ref{fig:methodology} summarizes the audit gates.

\subsection{Causal Ablation and an Oracle Headroom Estimate}
\label{subsec:oracle}

For each query state, we replay all $N$ candidates under a fixed
continuation policy and let a hindsight oracle select a successful one.
This estimates outcome-level selection headroom in the current candidate
pool; because it uses unavailable outcome information, it is not deployable.
We also probe input importance by leave-one-out patch masking, trusting the
probe only if irrelevant background ranks below relevant end-effector and
contact regions in a pre-registered sanity check.

\subsection{Pre-Registration and Statistical Power Control}
\label{subsec:prereg}

For the selection audit, we fix decision thresholds and the
replicate-consistency gate before inspecting the corresponding data and
report the gate's verdict even when it fails. A power gate avoids launching
experiments unlikely to yield a verdict: from a smoke sample, we estimate
the paired-residual standard deviation $\hat\sigma_r$ and project the
two-sided confidence-interval half-width at the planned sample size,
\begin{equation}
h(N) = z_{0.975}\,\hat\sigma_r / \sqrt{N},
\label{eq:power-gate}
\end{equation}
launching the full run only if $h(N_{\text{full}}) < \Delta_{\text{prereg}}$,
the separation between the pre-registered outcomes; otherwise the run is not
launched. Every reported
comparison states its bootstrap resampling unit and interval, so that an
effect whose significance depends on that choice is presented as such
rather than as a settled result. Closed-loop success is measured by
snapshot replay, in which we restore a state, execute a candidate action
chunk, and continue under a fixed dense base policy, rather than
inferred from a single-step action deviation.

\subsection{Scope}
\label{subsec:scope}

Our empirical findings are limited to Cosmos Policy on RoboCasa.
The audit components—premise checks, oracle headroom, a power gate, and
render-integrity verification—can also inform audits of other diffusion
control policies. We release a framebuffer-sentinel detector, a fix for
the identified OpenGL-context bug, and reproduction scripts; the detector
principle may transfer across backends, but this fix is stack-specific.

\begin{figure}[t]
\centering
\includegraphics[width=0.95\linewidth]{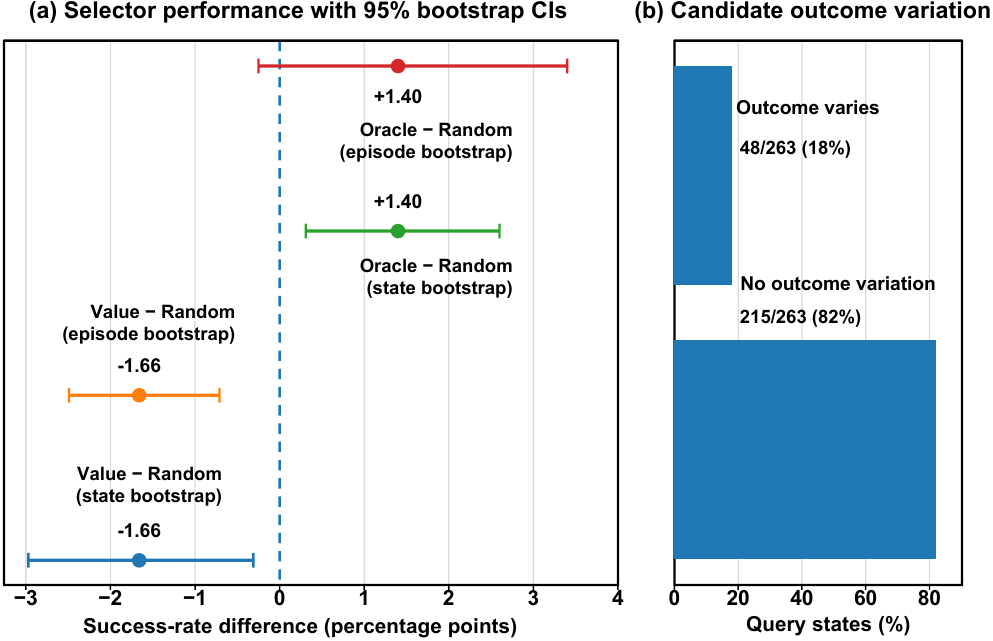}
\caption{Auditing best-of-$N$ planning selection. (a) Success-rate difference against the random baseline for the deployable value selector and the hindsight oracle, under state-level and episode-level bootstrap, with 95\% CIs. The value selector is below zero under both schemes; the oracle estimate crosses zero under episode-level resampling and is an upper bound, not a deployable selector. (b) Only a minority of query states show any outcome variation across candidates and repeated executions, leaving little for a selector to recover in the rest; a stricter decomposition (Supplement Section~\ref{supp:diversity}) further separates stable between-candidate differences from variation that is not stable across repeats.}
\label{fig:planning-audit}
\end{figure}

\section{Auditing Best-of-\texorpdfstring{$N$}{N} Planning Selection}
\label{sec:planning-audit}

We apply the audit to best-of-$N$ planning, comparing the model's
value-based ordering with random selection and estimating outcome-level
headroom using a hindsight oracle.

\subsection{Setup}
\label{subsec:planning-setup}

On render-verified clean data, we compare the value selector, random
selection, and the deployment-conditioned hindsight oracle on paired
sets of $N{=}8$ candidates with common continuation seeds. Differences
from random are reported with state-within-task and episode-within-task
bootstrap intervals.

\noindent\textbf{Continuation protocol.}
In snapshot replay, each candidate's first $16$ steps of a generated
$32$-step chunk are executed, stopping on success; the episode then replans
every $16$ steps under a fixed base policy until success or the $500$-step
task limit. The base policy uses one sample, all tokens, $10$ denoising steps,
and no value head; all selectors share the same outcome matrix and
continuation seeds ($R=2$). This controls the candidate's single-switch
effect but does not reproduce the full deployment distribution. The selection
audit's $10$-step continuation is distinct from the $2$-versus-$5$-step
allocation factorial, whose dense reference is $D(5,F)$.

\subsection{Weak Selection Signal and a Negative Value Selector}
\label{subsec:weak-signal}

The value selector is $1.66$\,pp worse than random (95\% CIs: state
bootstrap $[-2.97,-0.31]$; episode bootstrap $[-2.49,-0.71]$), driven
largely by TurnOnStove ($-3.87$\,pp). Excluding that task, the state-weighted
estimate remains negative ($-0.77$\,pp; episode CI $[-1.27,-0.21]$; state
CI $[-1.54,-0.03]$). Thus, the result is not uniform across tasks, but no
task reverses into a measured value-selector benefit.

The hindsight oracle's gain is $+1.40$\,pp, with a state-bootstrap CI of
$[+0.31,+2.60]$\,pp and an episode-bootstrap CI of $[-0.25,+3.40]$\,pp.
Under the pre-registered rule, the missed replicate-consistency gate and
episode-level interval make this weak, inconclusive evidence. With $R=2$,
the reliability estimate is only about $0.33$ (Supplement
Section~\ref{supp:stat}); we therefore do not infer a definitive ceiling
from either the cross-fitted estimate or the in-sample $+7.1$\,pp maximum.
More structurally, outcomes vary across candidates and/or repeats in only
$48/263$ states ($18\%$; Fig.~\ref{fig:planning-audit}b). The within-state
ICC is $0.20$, which indicates limited repeat agreement under this protocol;
the decomposition below distinguishes stable separation from variation
not stable across repeats.

\noindent\textbf{How much of this separability is real?}
Outcome uniformity is not explained simply by near-identical actions:
states with no outcome variation have no smaller action-space dispersion
(Supplement Section~\ref{supp:diversity}), though geometric diversity need not
be task-relevant. Of the $48$ varying states, $20$ show stable separation
(one candidate succeeds in both repeats and another fails in both); the other
$28$ do not meet this criterion. A label-shuffle null expects about $14.8$
stable-separable states, so the observed $20$ are not $20$ confirmed
opportunities. The descriptive oracle contribution is $+1.85$\,pp for the
first group and $-0.44$\,pp for the second; the pooled oracle gain under the
null is $+0.01$\,pp. With only two repeats, reproducible selection headroom
remains uncertain.

\subsection{Interpretation}
\label{subsec:interp}

Under this configuration, the negative value-selector result is clear,
whereas reproducible oracle headroom remains uncertain. The oracle measures
outcome-level headroom only for the observed candidates and specified
single-switch continuation, not general decision relevance. We therefore
make no claim that planning is universally ineffective; alternative
candidate-generation and continuation policies remain untested.

\begin{figure}[t]
\centering
\includegraphics[width=0.99\linewidth]{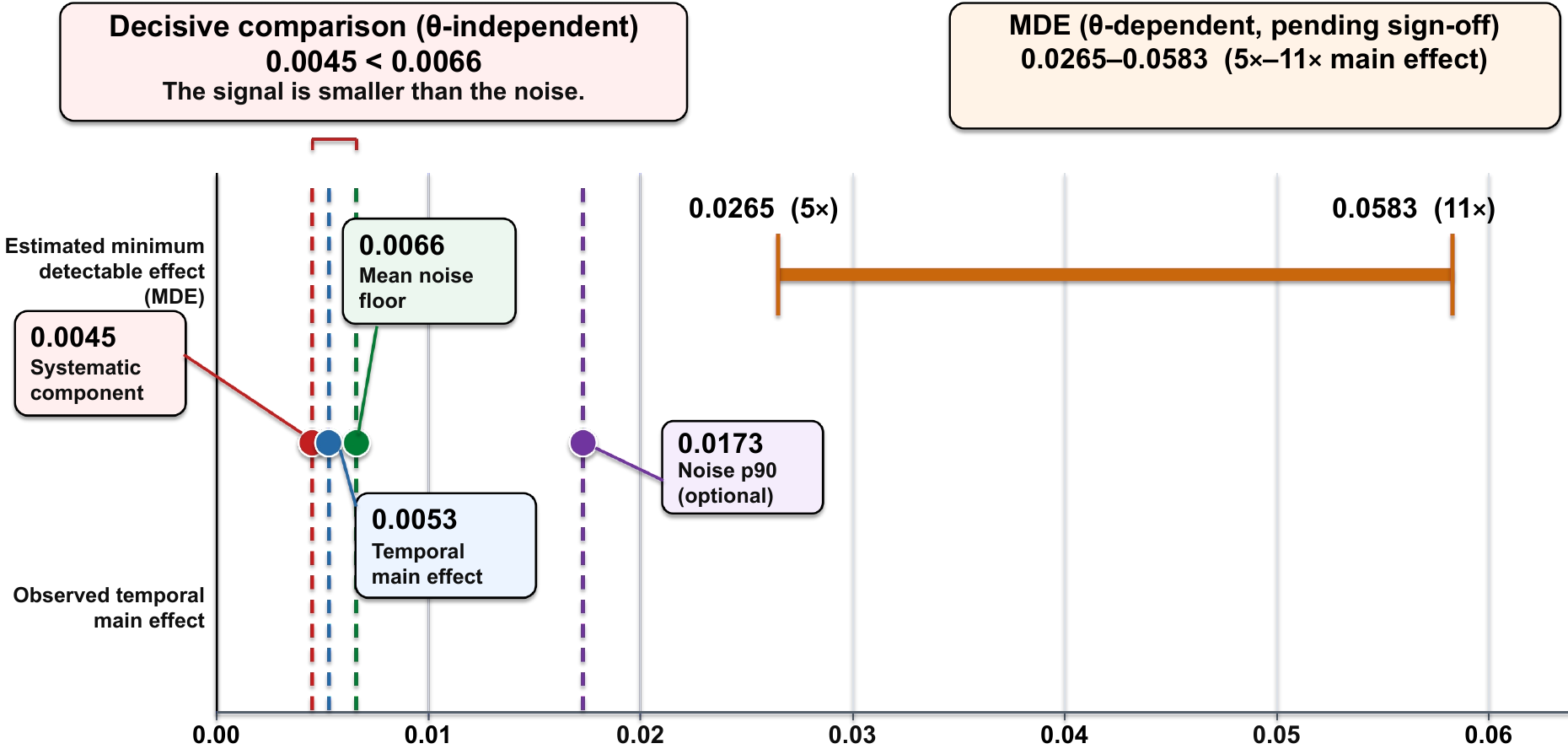}
\caption{Detectability of the joint interaction effect. The decisive comparison is threshold-free: the preliminary systematic component ($0.0045$) is below the mean per-state noise floor ($0.0066$) and comparable to the temporal main effect ($0.0053$) (noise $90$th percentile $0.0173$), so the effect cannot be resolved at any threshold. Under the locked threshold $\theta_B$ the minimum detectable effect is $0.0265$ to $0.0583$ ($5$ to $11$ times the main effect), shown for reference. Values are from a $16$-state smoke sample whose preregistered power gate was not cleared, so the full factorial was not launched.}
\label{fig:mde}
\end{figure}

\begin{table*}[t]
\centering
\footnotesize
\setlength{\tabcolsep}{5pt}
\caption{Measured per-forward FLOPs (a), measured end-to-end query speedup under
whole-frame token pruning (b), and the measured dense-query component breakdown
(c), on Cosmos Policy (RoboCasa, batch 1, 5 denoising steps, 11 latent frames
$\times$ 196 tokens $=$ 2156; NVIDIA RTX A6000, bf16, eager, $N=30$). Panel (b)
directly times the complete query, dense against pruned, not a composition of
component times: the deployed path is the query as the eval calls it
(future-image decode and value extraction included); the decode-excluded path
removes future-image decode on both sides and is most favorable to pruning. This
is a latency-only measurement (dropped frames receive zero output), so action
quality is not assessed; no pruning level on either path reaches the
pre-registered $1.5\times$ target. Setup, environment and result-file paths are
in Supplement Section~\ref{supp:timing}.}
\label{tab:timing-breakdown}

\begin{minipage}{\textwidth}
\centering
(a) Per-forward compute (FLOPs, measured via FlopCounterMode: matmul $+$ SDPA)\\[2pt]
\begin{tabular}{lcccc}
\toprule
Pruning level & Tokens & TFLOP & Nominal saving & of which SDPA (TFLOP) \\
\midrule
Dense                         & 2156 & 8.535 & ---    & 1.319 \\
Drop 1 future image           & 1960 & 7.682 & 10.0\% & 1.111 \\
Drop 2 future images          & 1764 & 6.846 & 19.8\% & 0.921 \\
All future images             & 1568 & 6.028 & 29.4\% & 0.748 \\
All future frames             & 1372 & 5.228 & 38.7\% & 0.593 \\
All future frames $+$ value   & 1176 & 4.446 & 47.9\% & 0.455 \\
\bottomrule
\end{tabular}
\end{minipage}

\vspace{5pt}

\begin{minipage}{0.64\textwidth}
\centering
(b) End-to-end query speedup (ms / $\times$, measured, $N=30$)\\[2pt]
\begin{tabular}{lccc}
\toprule
Pruning level & Deployed path & Decode-excluded & $\geq 1.5\times$? \\
\midrule
Dense                       & 1254.9 / ---   & \phantom{0}812.7 / ---   & --- \\
Drop 1 future image         & 1232.1 / 1.019 & \phantom{0}791.2 / 1.027 & no \\
Drop 2 future images        & 1171.0 / 1.072 & \phantom{0}728.9 / 1.115 & no \\
All future images           & 1140.5 / 1.100 & \phantom{0}699.8 / 1.161 & no \\
All future frames           & 1096.5 / 1.144 & \phantom{0}654.0 / 1.243 & no \\
All future frames $+$ value & 1088.9 / \textbf{1.152} & \phantom{0}648.6 / \textbf{1.253} & no \\
\bottomrule
\end{tabular}
\end{minipage}
\hfill
\begin{minipage}{0.34\textwidth}
\centering
(c) Dense component (ms, measured)\\[2pt]
\begin{tabular}{lc}
\toprule
Component (per query) & ms \\
\midrule
VAE encode ($1\times$)          & 238.0 \\
DiT forward ($5\times$, summed) & 530.7 \\
Future-image VAE decode         & 434.4 \\
\midrule
Full query                      & 1241.8 \\
\bottomrule
\end{tabular}
\end{minipage}
\end{table*}

\section{A Feasibility Audit of Adaptive Compute Allocation}
\label{sec:allocation}

We audit whether reducing denoising steps or processing fewer tokens
can deliver practical efficiency on this stack. Because the preregistered
criteria are end-to-end latency and a resolvable joint interaction effect,
and the power gate was not cleared, this is a feasibility diagnostic rather
than a confirmatory test of joint allocation.

\subsection{Nominal FLOPs Savings Do Not Convert to Wall-Clock Savings}
\label{subsec:flops-clock}

Dropping all future frames plus the value frame removes $47.9\%$ of
per-forward FLOPs, but direct complete-query timing shows only $1.152\times$
speedup on the deployed eager path and $1.253\times$ when future-image decode
is excluded; neither reaches the preregistered $1.5\times$ target
(Table~\ref{tab:timing-breakdown}). This fixed whole-frame mask adds no
per-token scoring. The five DiT forwards account for $530.7$ ms of the
$1241.8$ ms dense query, while VAE encode ($238.0$ ms) and future-image decode
($434.4$ ms) are not reduced by token pruning. This is a latency-only result
(dropped frames receive zero output); full settings and per-level latency
percentiles are in Supplement Section~\ref{supp:timing} and Table~\ref{tab:e2e-detail}.

\subsection{The Preliminary Interaction Effect Is Below the Noise Resolution}
\label{subsec:detection-limit}

We then ask whether a joint-allocation benefit could be resolved on this
stack even if present. Using the pre-registered factorial diagnostic of
Section~\ref{sec:methodology}, crossing two denoising schedules with full, key-token, and
uniform-token ($U$) selection, the joint-allocation effect is the
uniform-token interaction
\begin{equation}
I_U = D(2,U) - D(2,F) - D(5,U) + D(5,F),
\qquad \mathbb{E}[I_U] = 0
\label{eq:interaction-main}
\end{equation}
under exact additivity, where $D(k,s)$ is the paired closed-loop success at
denoising schedule $k$ under selection $s$, relative to the baseline
$D(5,F) = 0$. Because seed noise largely cancels in this
difference-in-differences structure, with a measured effective
correlation of $\rho = 0.92$, the residual noise is small but
structured: per-state noise magnitude has mean $0.0066$ and $90$th
percentile $0.0173$. The systematic component, estimated after
noise-bias removal from a $16$-state smoke sample, is about $0.0045$. 

The decisive comparison is independent of any chosen threshold: the
preliminary systematic component ($0.0045$) is smaller than the mean
noise floor ($0.0066$) and comparable to the temporal main effect
itself ($0.0053$) in the same states. A joint effect that is smaller
than the per-state noise it must be read against cannot be resolved from
this measurement, whatever decision threshold is adopted.
Figure~\ref{fig:mde} places these quantities on the same absolute scale.
For completeness, under the currently locked threshold $\theta_B$
(Supplement Eq.~\ref{eq:theta}) the
minimum detectable interaction is $0.0265$ to $0.0583$, that is $5$ to
$11$ times the temporal main effect; we note that this range depends on
the threshold calibration still pending sign-off, and it will be
recomputed for the camera-ready version, but the threshold-free
comparison above does not depend on it.

This is inconclusive, not a null result: the full factorial was not
launched because the power gate failed. The estimated sample requirement—
hundreds to roughly two thousand states, is an extrapolation from the current
detection limit (Supplement Section~\ref{supp:stat}), not a verified power
result. Whether joint allocation helps on this stack remains open.

\subsection{Implications}
\label{subsec:allocation-implications}

Efficiency claims should pair FLOPs with complete-query latency, and
allocation claims should be judged against a preregistered detection limit.
Here, token-pruning savings miss the latency target and the smoke-scale joint
effect is unresolved; this does not establish that adaptive allocation is
ineffective in general.

\section{Evaluation Reliability: A Silent Rendering Corruption}
\label{sec:reliability}

We identify a third threat to test-time-scaling evaluation: silent,
compute-dependent rendering corruption in the RoboCasa/\texttt{robosuite}
stack, to which compute-intensive inference paths are more exposed.

\subsection{Mechanism}
\label{subsec:mechanism}

The corruption originates in OpenGL context management during
environment resets. On a hard reset the stack constructs a new
simulation instance while an orphaned instance remains eligible for
garbage collection; when the orphan is later collected, its GL context
is freed while a different context is active, silently invalidating the
framebuffer of the live environment. Subsequent renders return corrupted
observations, frozen or noise-like frames, with no error raised. Figure~\ref{fig:corruption} contrasts a corrupted frame with the correct render of the same state. We confirm the mechanism with a framebuffer sentinel that pre-fills each
render target with a marker and flags targets left substantially
untouched after rendering.\footnote{We attribute this corruption to OpenGL context management during hard resets in the tested stack, confirmed by our framebuffer sentinel and eliminated by our fix. We have not sought maintainer confirmation of the root-cause attribution.}

\subsection{Effect on the Official Direct-Policy Benchmark}
\label{subsec:official-impact}

We first ask whether this corruption materially affects the officially
reported RoboCasa numbers. We run the unmodified official evaluation
file in two arms, bug-present and render-fixed, paired episode by
episode on the four tasks where our baseline run exhibited corruption,
over 50 consecutive episodes each (200 paired episodes), with the
framebuffer sentinel active throughout.

\begin{table}[t]
\centering
\small
\begin{tabular}{@{}lccc@{}}
\toprule
Task & Succ (bug$\rightarrow$fix) & Corrupt & Flips \\
\midrule
PnPCounterToMicrowave & $16 \rightarrow 18$ & 3 & $+2$ \\
PnPMicrowaveToCounter & $18 \rightarrow 18$ & 3 & $0$ \\
PnPStoveToCounter     & $32 \rightarrow 33$ & 1 & $+1$ \\
TurnSinkSpout         & $36 \rightarrow 36$ & 1 & $0$ \\
\midrule
\textbf{Total}        & $\mathbf{102 \rightarrow 105}$ & $8/200$ & $+3$ \\
\bottomrule
\end{tabular}
\caption{Paired bug-vs-fix comparison under the official direct-policy
protocol. All three outcome flips are corrupted episodes moving from
failure to success; none move the other way. Among the 192 uncorrupted
episodes, success is identical in both arms.}
\label{tab:official-impact}
\end{table}

As Table~\ref{tab:official-impact} shows, under the direct-policy
protocol the measured corruption rate is $8/1200$ ($0.67\%$ per episode
in the bug arm, and zero in the fixed arm), and the paired effect of
fixing it is $+3$ successes over 200 episodes. All flips move corrupted
episodes from failure to success, so the corruption understates success.
Extrapolated to the full 24-task, 50-episode protocol, the effect is
about $+0.25$\,pp and never exceeds an upper bound of $+0.6$\,pp. We
conclude that the officially reported direct-policy figure of $67.1\%$ is
not materially affected: the corruption is real and present in the
official stack, but its direct-policy impact is small and in the
direction of underreporting success.

\subsection{Corruption Scales with Per-Inference Compute}
\label{subsec:scaling}

The direct-policy result does not bound the risk for test-time scaling,
because corruption frequency depends on how much computation each query
allocates. We observe a sharp, monotone dependence:

\begin{itemize}
\item \textbf{Direct policy} (one forward pass per query): $0.67\%$ of
episodes corrupted, paired-verified in
Table~\ref{tab:official-impact}.
\item \textbf{Direct-mode best-of-$N$} ($N{=}8$ candidates, simple
forward passes): $0/40$ episodes corrupted, with identical success
($29/40$) in both arms. Additional sampling alone does not raise the
rate.
\item \textbf{Planning mode} (separate future-state and value calls per
query, substantially more allocation per query): up to $65\%$ of
episodes corrupted.
\end{itemize}

The $65\%$ figure is observed under a compute-heavy planning
configuration and should be read as an upper-bound illustration of the
trend, not as the corruption rate of any standard benchmark. RoboCasa
provides no native planning-mode evaluation, and this configuration is
non-standard; the official repository exposes best-of-$N$ planning only
for a real-robot platform that does not pass through \texttt{robosuite},
so the planning-path corruption touches no number the official
benchmark reports. On the LIBERO stack, which on a hard reset explicitly
destroys the prior simulation before constructing a new one and so
leaves no orphan, the trigger condition is not met; we infer from the
source that the reported LIBERO figure is unaffected, though we did not
run that environment.

\subsection{Implications}
\label{subsec:reliability-implications}

The official direct-policy result is largely unaffected, but compute-heavy
planning is more exposed. Studies using this stack should verify render
integrity before attributing gains to inference-time methods.

\section{Discussion}
\label{sec:discussion}

Our three audits are independent, but they point in the same direction:
on the tested Cosmos Policy configuration, the computation added at
inference time to realize test-time scaling does not buy what it is
assumed to buy, and part of the apparent benefit reported in this
setting can arise from evaluation artifacts rather than from the method.

\noindent\textbf{What the findings do and do not say.} None of the three
results claims that planning, adaptive allocation, or test-time scaling is
useless in general; each is stated at the scope of the tested
configuration. The planning result is that the deployable selector is
worse than random and that selection headroom is small, not that no
selector can ever help. The allocation result is that nominal FLOPs
savings do not convert to end-to-end speedup here, together with a
detectability statement from a smoke-scale sample whose preregistered
power gate was not cleared, not a confirmed null. The reliability result
is that a real, compute-dependent corruption exists in the official stack
while leaving the reported direct-policy number essentially intact.

\noindent\textbf{A methodology, not only a set of negative results.} The value
of a negative audit depends on whether its measurements can be trusted,
which is why the diagnostic methodology is our primary contribution.
Premise-first verification keeps us from building on a signal that does
not exist; the oracle bounds the achievable gain before any selector is
tuned; the power gate stops an experiment that cannot yield a verdict
before it consumes compute; and render-integrity checking protects every
closed-loop number. These components are model-agnostic and transfer to
any audit of test-time compute in a diffusion-based control policy, even
though the specific render fix is tied to the stack we study. We release
the detector, fix, scripts, and preregistered protocols so the audits can
be repeated. Our strongest limitation is scope: whether these findings
generalize to other models, simulators, or real hardware is an open
question the methodology is designed to answer rather than presuppose, and
we see the audits not as a verdict against inference-time computation but
as checks that should precede any claim that test-time scaling improves
control.

\section{Conclusion}
\label{sec:conclusion}

We presented an empirical audit of test-time scaling in a representative
world-action model, Cosmos Policy on RoboCasa, built around a
premise-first, pre-registered methodology with explicit statistical
power control and render-integrity verification. The audit yields three
independent reasons an apparent gain may not be real: best-of-$N$
candidate selection carries little usable signal and its deployable
selector is worse than random; nominal compute savings from token-level
allocation do not convert to practical speedup, and the joint allocation
effect lies below the system's noise resolution; and a silent,
compute-dependent rendering corruption inflates the reliability risk of
exactly the compute-intensive inference paths that test-time scaling
relies on, while leaving the official direct-policy benchmark intact. We
release the methodology and tools so that these checks can be applied
before performance gains are attributed to test-time scaling rather than
to unverified assumptions or evaluation artifacts.

Furthermore, we will investigate advanced techniques such as model pruning~\cite{ji2025computation,li2025mutual,liu2025toward,Zhang_2026_CVPR}, which enable neural networks to maintain computational efficiency as the number of parameters increases, particularly in resource-limited settings~\cite{liu2025rora,niu2025mobile,tan2025zeroqat,zhang2025towards,10.1145/3747842,tan2025perturbation,yang2025fairsmoe,yuan2021work,yuan2022mobile,liu2023scalable,liu-etal-2026-rcr,liu2026should,liu2026structured}. we will employ diffusion-based generative models~\cite{meng2024instructgie} to synthesize compact, informative samples~\cite{liu2023interpretable,liu2021explainable,liu2022efficient}, thereby facilitating scalable experimentation in healthcare-related tasks~\cite{liu2024brain,wang2026learning,wang2025fairgnn,chinta2023optimization,wang2025graph,yin2026assisting,wang2025amcr,chinta2025ai}. These synthetic samples will be integrated with large language models (LLMs) agents~\cite{xu2026lever,wang2026parprogressalignedrecurrencelength,zhang-etal-2026-memmark,wang-etal-2026-memguard,dong2026conditionalrankallocationtaxonomyaware} to support downstream applications such as 3D reconstruction~\cite{dong2026imageslookrightretrieve,kang2026drilltrainingworldmodels,li2025local,dong2024physical,dong2024df}, enhancing overall model utility in multi-modal scenarios~\cite{liu2026dualpathwaycircuitsobjecthallucination,zhang-etal-2026-modality,dong2026bridginglowlevelimagefusion,dong2026lance,dong2026closeloopretrievabilitygap}.


{
    \small
    \bibliographystyle{ieeenat_fullname}
    \bibliography{main}

@String(CVPR= {IEEE Conf. Comput. Vis. Pattern Recog.})

@String(ICCV= {Int. Conf. Comput. Vis.})

@String(ICASSP=	{ICASSP})

@String(ICLR = {Int. Conf. Learn. Represent.})

@String(AAAI = {AAAI})

@String(CVPR  = {CVPR})

@String(ICCV  = {ICCV})

@String(ICLR  = {ICLR})

@inproceedings{
kim2026cosmos,
title={Cosmos Policy: Fine-Tuning Video Models for Visuomotor Control and Planning},
author={Moo Jin Kim and Yihuai Gao and Tsung-Yi Lin and Yen-Chen Lin and Yunhao Ge and Grace Lam and Percy Liang and Shuran Song and Ming-Yu Liu and Chelsea Finn and Jinwei Gu},
booktitle={The Fourteenth International Conference on Learning Representations},
year={2026},
url={https://openreview.net/forum?id=wPEIStHxYH}
}

@inproceedings{zhu2025irasim,
  title={Irasim: A fine-grained world model for robot manipulation},
  author={Zhu, Fangqi and Wu, Hongtao and Guo, Song and Liu, Yuxiao and Cheang, Chilam and Kong, Tao},
  booktitle={2025 IEEE/CVF International Conference on Computer Vision (ICCV)},
  pages={9834--9844},
  year={2025},
  organization={IEEE}
}

@inproceedings{3780338.3782852,
author = {Wang, Zhendong and Li, Zhaoshuo and Mandlekar, Ajay and Xu, Zhenjia and Fan, Jiaojiao and Narang, Yashraj and Fan, Linxi and Zhu, Yuke and Balaji, Yogesh and Zhou, Mingyuan and Liu, Ming-Yu and Zeng, Yu},
title = {One-step diffusion policy: fast visuomotor policies via diffusion distillation},
year = {2025},
publisher = {JMLR.org},
booktitle = {Proceedings of the 42nd International Conference on Machine Learning},
articleno = {2514},
numpages = {18},
location = {Vancouver, Canada},
series = {ICML'25}
}

@inproceedings{ji2026bac,
  title     = {{Block-wise Adaptive Caching for Accelerating
               Diffusion Policy}},
  author    = {Ji, Kangye and Meng, Yuan and Cui, Hanyun
               and Li, Ye and Zhou, Jianbo and Hua, Shengjia
               and Chen, Lei and Wang, Zhi},
  booktitle = {International Conference on Learning
               Representations (ICLR)},
  year      = {2026},
  url       = {https://proceedings.iclr.cc/paper_files/paper/2026/hash/2712b17bb58ea5b2b65c45857b024744-Abstract-Conference.html}
}

@InProceedings{ji2026sag,
  title     = {{Sparse ActionGen: Accelerating Diffusion Policy
               with Real-time Pruning}},
  author    = {Ji, Kangye and Zhou, Jianbo and Meng, Yuan
               and Li, Ye and Cui, Hanyun and Wang, Zhi},
  booktitle = {Proceedings of the 43rd International
               Conference on Machine Learning},
  volume    = {306},
  series    = {Proceedings of Machine Learning Research},
  pages     = {51692--51711},
  year      = {2026},
  publisher = {PMLR},
  url       = {https://proceedings.mlr.press/v306/ji26o.html}
}

@inproceedings{robocasa365,
  title={RoboCasa365: A Large-Scale Simulation Framework for Training and Benchmarking Generalist Robots},
  author={Soroush Nasiriany and Sepehr Nasiriany and Abhiram Maddukuri and Yuke Zhu},
  booktitle={International Conference on Learning Representations (ICLR)},
  year={2026}
}

@inproceedings{
li2024evaluating,
title={Evaluating Real-World Robot Manipulation Policies in Simulation},
author={Xuanlin Li and Kyle Hsu and Jiayuan Gu and Oier Mees and Karl Pertsch and Homer Rich Walke and Chuyuan Fu and Ishikaa Lunawat and Isabel Sieh and Sean Kirmani and Sergey Levine and Jiajun Wu and Chelsea Finn and Hao Su and Quan Vuong and Ted Xiao},
booktitle={8th Annual Conference on Robot Learning},
year={2024},
url={https://openreview.net/forum?id=LZh48DTg71}
}

@inproceedings{jang2026verifier,
  title={Verifier-free test-time sampling for vision-language-action models},
  author={Jang, Suhyeok and Kim, Dongyoung and Kim, Changyeon and Kim, Youngsuk and Shin, Jinwoo},
  booktitle={International Conference on Learning Representations},
  volume={2026},
  pages={139997--140017},
  year={2026}
}

@inproceedings{
zhang2026worldinworld,
title={World-In-World: World Models in a Closed-Loop World},
author={Jiahan Zhang and Muqing Jiang and Nanru Dai and TaiMing Lu and Arda Uzunoglu and Shunchi Zhang and Yana Wei and Jiahao Wang and Vishal M. Patel and Paul Pu Liang and Daniel Khashabi and Cheng Peng and Rama Chellappa and Tianmin Shu and Alan Yuille and Yilun Du and Jieneng Chen},
booktitle={The Fourteenth International Conference on Learning Representations},
year={2026},
url={https://openreview.net/forum?id=yDmb7xAfeb}
}

@inproceedings{
li2026vlaattc,
title={{VLA}-{ATTC}: Adaptive Test-Time Compute for {VLA} Models with Relative Action Critic Model},
author={Wenhao Li and Xiu Su and Yichao Cao and Hongyan Xu and Xiaobo Xia and Shan You and Yi Chen and Chang Xu},
booktitle={Forty-third International Conference on Machine Learning},
year={2026},
url={https://openreview.net/forum?id=tAqHcMkhy7}
}

@inproceedings{
bilal2026what,
title={What If We Allocate Test-Time Compute Adaptively?},
author={Ahsan Bilal and Muhammad Ahmed Mohsin and Muhammad Umer and Ali Subhan and Hassan Rizwan and Ayesha Mohsin and Dean F. Hougen},
booktitle={Forty-third International Conference on Machine Learning},
year={2026},
url={https://openreview.net/forum?id=wblvinAlrW}
}

@inproceedings{
chen2026manipevalagent,
title={ManipEvalAgent: Promptable and Efficient Evaluation Framework for Robotic Manipulation Policies},
author={Yiteng Chen and Huiping Zhuang and WenboLi and Shiyi Wang and Xiangyu Zhao and Qingyao Wu},
booktitle={The Fourteenth International Conference on Learning Representations},
year={2026},
url={https://openreview.net/forum?id=3u6AkbWEls}
}

@inproceedings{
jangir2026robotarena,
title={RobotArena \${\textbackslash}infty\$: Scalable Robot Benchmarking via Real-to-Sim Translation},
author={Yash Jangir and Yidi Zhang and Kashu Yamazaki and Chenyu Zhang and Kuan-Hsun Tu and Tsung-Wei Ke and Lei Ke and Yonatan Bisk and Katerina Fragkiadaki},
booktitle={The Fourteenth International Conference on Learning Representations},
year={2026},
url={https://openreview.net/forum?id=OutljIofvS}
}

@inproceedings{
zhao2025sample,
title={Sample, Scrutinize and Scale: Effective Inference-Time Search by Scaling Verification},
author={Eric Zhao and Pranjal Awasthi and Sreenivas Gollapudi},
booktitle={Forty-second International Conference on Machine Learning},
year={2025},
url={https://openreview.net/forum?id=wl3eI4wiE5}
}

@inproceedings{
hua2026learning,
title={Learning to Explore: Scaling Agentic Reasoning via Exploration-Aware Policy Optimization},
author={Xingyuan Hua and Sheng Yue and Ju Ren},
booktitle={Forty-third International Conference on Machine Learning},
year={2026},
url={https://openreview.net/forum?id=YNPHJ2A2oV}
}

@inproceedings{shum2026swe,
  title={SWE-RM: Execution-free Feedback for Software Engineering Agents},
  author={Shum, KaShun and Hui, Binyuan and Chen, Jiawei and Zhang, Lei and Yang, Jiaxi and Huang, Yuzhen and Lin, Junyang and He, Junxian and others},
  booktitle={International Conference on Learning Representations},
  volume={2026},
  pages={143426--143446},
  year={2026}
}

@inproceedings{
li2026scaling,
title={Scaling Robot Policy Evaluation via Discrete Diffusion World Model},
author={Yaxuan Li and Junjie Wen and Zhongyi Zhou and Yefei Chen and Chaomin Shen and Yaxin Peng and Yichen Zhu},
booktitle={Forty-third International Conference on Machine Learning},
year={2026},
url={https://openreview.net/forum?id=93uNlQ1Qp0}
}

@inproceedings{
guo2024prediction,
title={Prediction with Action: Visual Policy Learning via Joint Denoising Process},
author={Yanjiang Guo and Yucheng Hu and Jianke Zhang and Yen-Jen Wang and Xiaoyu Chen and Chaochao Lu and Jianyu Chen},
booktitle={The Thirty-eighth Annual Conference on Neural Information Processing Systems},
year={2024},
url={https://openreview.net/forum?id=teVxVdy8R2}
}

@inproceedings{
alonso2024diffusion,
title={Diffusion for World Modeling: Visual Details Matter in Atari},
author={Eloi Alonso and Adam Jelley and Vincent Micheli and Anssi Kanervisto and Amos Storkey and Tim Pearce and Fran{\c{c}}ois Fleuret},
booktitle={The Thirty-eighth Annual Conference on Neural Information Processing Systems},
year={2024},
url={https://openreview.net/forum?id=NadTwTODgC}
}

@inproceedings{
li2026step,
title={{STEP}: Warm-Started Visuomotor Policies with Spatiotemporal Consistency Prediction},
author={Jinhao Li and Yuxuan Cong and Yingqiao Wang and Hao Xia and Shan Huang and Yijia Zhang and Ningyi Xu and Guohao Dai},
booktitle={Forty-third International Conference on Machine Learning},
year={2026},
url={https://openreview.net/forum?id=nWtjaAmug8}
}

@inproceedings{
chun2026sparse,
title={Sparse Imagination for Efficient Visual World Model Planning},
author={Junha Chun and Youngjoon Jeong and Taesup Kim},
booktitle={The Fourteenth International Conference on Learning Representations},
year={2026},
url={https://openreview.net/forum?id=faxcxKINBC}
}

@article{zhou2026learning,
  title={Learning 3d persistent embodied world models},
  author={Zhou, Siyuan and Du, Yilun and Yang, Yuncong and Han, Lei and Chen, Peihao and Yeung, Dit-Yan and Gan, Chuang},
  journal={Advances in Neural Information Processing Systems},
  volume={38},
  pages={104807--104829},
  year={2026}
}

@inproceedings{lu2025gwm,
  title={Gwm: Towards scalable gaussian world models for robotic manipulation},
  author={Lu, Guanxing and Jia, Baoxiong and Li, Puhao and Chen, Yixin and Wang, Ziwei and Tang, Yansong and Huang, Siyuan},
  booktitle={2025 IEEE/CVF International Conference on Computer Vision (ICCV)},
  pages={9263--9274},
  year={2025},
  organization={IEEE}
}

@inproceedings{
yin2026ddpwm,
title={{DDP}-{WM}: Disentangled Dynamics Prediction for Efficient World Models},
author={Shicheng Yin and Kaixuan Yin and Weixing Chen and Yang Liu and Guanbin Li and Liang Lin},
booktitle={Forty-third International Conference on Machine Learning},
year={2026},
url={https://openreview.net/forum?id=gEc0iJCYxw}
}

@inproceedings{
cohen2026horizon,
title={Horizon Imagination: Efficient On-Policy Rollout in Diffusion World Models},
author={Lior Cohen and Ofir Nabati and Kaixin Wang and Navdeep Kumar and Shie Mannor},
booktitle={The Fourteenth International Conference on Learning Representations},
year={2026},
url={https://openreview.net/forum?id=Obefq4k8iG}
}

@inproceedings{
kim2026dexterous,
title={Dexterous World Models},
author={Byungjun Kim and Taeksoo Kim and Junyoung Lee and Hanbyul Joo},
booktitle={Conference on Computer Vision and Pattern Recognition 2026},
year={2026},
url={https://openreview.net/forum?id=jLa7jmBrRc}
}

@article{guo2026xwam,
  title={Unified 4D World Action Modeling from Video Priors with Asynchronous Denoising},
  author={Guo, Jun and Li, Qiwei and Li, Peiyan and Chen, Zilong and Sun, Nan and Su, Yifei and Wang, Heyun and Zhang, Yuan and Li, Xinghang and Liu, Huaping},
  journal={arXiv preprint arXiv:2604.26694},
  year={2026}
}

@article{liu2026dualpathwaycircuitsobjecthallucination,
      title={Dual-Pathway Circuits of Object Hallucination in Vision-Language Models}, 
      author={Jiaxin Liu and Ding Zhong and Yue Wang and Zhidong Yang and Zhaolu Kang and Guangyuan Dong and Qishi Zhan and Pengcheng Fang and Aofan Liu},
      year={2026},
      eprint={2605.13156},
      archivePrefix={arXiv},
      primaryClass={cs.CV},
      url={https://arxiv.org/abs/2605.13156},
      journal={arXiv preprint arXiv:2605.13156},
      note={Accepted at NeurIPS 2026},
}

@inproceedings{zhang-etal-2026-modality,
    title = "Modality Disentangled Learning for Incomplete Multimodal Emotion Recognition: A Primitive Memory Distillation Perspective",
    author = {Zhang, Jiaqi  and
      Pang, Zheng  and
      Li, Mengting  and
      Wang, Yiqi  and
      Dong, Guangyuan  and
      Xue, Chao  and
      Wu, Yusen  and
      Li, Zihao  and
      Phan, Huy  and
      Zhao, Sicheng  and
      Schuller, Bj{\"o}rn  and
      Luo, Jiachen},
    booktitle = "Findings of the {A}ssociation for {C}omputational {L}inguistics: {EMNLP} 2026",
    month = oct,
    year = "2026",
    publisher = "Association for Computational Linguistics",
    url = "https://aclanthology.org/2026.findings-emnlp.2220/",
    doi = "10.18653/2026.findings-emnlp.2220",
    pages = "47534--47552",
}

@article{dong2026bridginglowlevelimagefusion,
      title={Bridging Low-Level Image Fusion and High-Level Semantic Understanding}, 
      author={Guangyuan Dong and Lingdong Shen and Xudong Zhang and Haodong Jing and Haoyu Wang and Zihao Li and Zonghao Ying and Ziyu Song and Shenghao Liu and Jiachen Luo and Hao Zhang},
      year={2026},
      eprint={2611.02896},
      archivePrefix={arXiv},
      primaryClass={cs.AI},
      url={https://arxiv.org/abs/2611.02896},
      journal={arXiv preprint arXiv:2611.02896},
}

@inproceedings{dong2026lance,
  title = {{Lance}: A {LANGUAGE-ANCHORED CLOSED-LOOP ESTIMATOR} for Bridging Low-Level Image Fusion and High-Level Semantic Understanding in Remote Sensing},
  author = {Dong, Guangyuan and Shen, Lingdong and Zhang, Xudong and Ying, Zonghao and Jing, Haodong and Song, Ziyu and Liu, Shenghao and Luo, Jiachen and Zhang, Hao and Wang, Haoyu and Li, Zihao},
  booktitle = {Proceedings of the 35th ACM International Conference on Information and Knowledge Management},
  year = {2026},
  pages = {1036--1047},
  publisher = {Association for Computing Machinery},
  doi = {10.1145/3799682.3841149},
  url = {https://doi.org/10.1145/3799682.3841149}
}

@article{dong2026closeloopretrievabilitygap,
      title={Close the Loop on the Retrievability Gap}, 
      author={Guangyuan Dong and Wenhao Li and Zhen Tian and Jiaqi Zhang and Haoyu Wang and Junhao Xu and Mingzhe Liu and Pan Wang and Fei Shen and Tat-Seng Chua},
      year={2026},
      eprint={2611.28146},
      archivePrefix={arXiv},
      primaryClass={cs.CV},
      url={https://arxiv.org/abs/2611.28146},
      journal={arXiv preprint arXiv:2611.28146},
}

@inproceedings{zhang-etal-2026-memmark,
    title = "{MEMMARK}: State-Evolution Attribution Watermarking for Agent Long-Term Memory Systems",
    author = "Zhang, Haobo  and
      Mao, Xutao  and
      Dong, Guangyuan  and
      Li, Ziwei  and
      Su, Xuanbo  and
      Chen, Kaijie  and
      Yang, Jing  and
      Lin, Zheng",
    booktitle = "Findings of the {A}ssociation for {C}omputational {L}inguistics: {EMNLP} 2026",
    month = oct,
    year = "2026",
    publisher = "Association for Computational Linguistics",
    url = "https://aclanthology.org/2026.findings-emnlp.395/",
    doi = "10.18653/2026.findings-emnlp.395",
    pages = "8333--8349",
}

@inproceedings{wang-etal-2026-memguard,
    title = "{M}em{G}uard: Persisting Verifier Signals for {LLM}-Agent Memory Governance",
    author = "Wang, Haoyu  and
      Dong, Guangyuan  and
      He, Liang  and
      Zhang, Zijing  and
      Luo, Jiachen  and
      Xue, Chao  and
      Liu, Chuang  and
      Tang, Hao",
    booktitle = "Proceedings of the 2026 Conference on Empirical Methods in Natural Language Processing",
    month = oct,
    year = "2026",
    publisher = "Association for Computational Linguistics",
    url = "https://aclanthology.org/2026.emnlp-main.423/",
    doi = "10.18653/2026.emnlp-main.423",
    pages = "9677--9708",
}

@article{wang2026parprogressalignedrecurrencelength,
      title={{PAR}: Progress-Aligned Recurrence for Length Generalization in State Tracking}, 
      author={Pan Wang and Yihao Hu and Guangyuan Dong and Yitong Lin and Hang Wang and Zimi Liu and Zhihao Wen and Jianshe Li and Jiang-Ming Yang and Wei Wu and Yongqi Tong},
      year={2026},
      eprint={2610.27628},
      archivePrefix={arXiv},
      primaryClass={cs.CL},
      url={https://arxiv.org/abs/2610.27628},
      journal={arXiv preprint arXiv:2610.27628},
}

@article{dong2026conditionalrankallocationtaxonomyaware,
      title={Conditional Rank Allocation for Taxonomy-Aware Medical Language Model Adaptation}, 
      author={Guangyuan Dong and Ziwei Hong and Xuehao Zhou and Zidong Yu and Bingchen Liu and Kehan Liu and Chuang Liu and Rong Fu and Yuchao Hou},
      year={2026},
      eprint={2610.06765},
      archivePrefix={arXiv},
      primaryClass={cs.AI},
      url={https://arxiv.org/abs/2610.06765},
      journal={arXiv preprint arXiv:2610.06765},
      note={Accepted at IEEE BIBM 2026},
}

@article{dong2026imageslookrightretrieve,
      title={When Images Look Right and Retrieve Wrong: Coverage-Guided Cross-Scale Re-Indexing for Knowledge-Faithful Generative Perception}, 
      author={Guangyuan Dong and Chuang Liu and Haoyu Wang and Yangchen Zeng and Jiaqi Zhang and Li Jiuxing and Xiaoyang Yu and Pinlong Zhao and Yuchao Hou and Ziwei Li and Zheng Lin and Alexander Lim Han Yang and Yusen Wu},
      year={2026},
      eprint={2608.20810},
      archivePrefix={arXiv},
      primaryClass={cs.MM},
      url={https://arxiv.org/abs/2608.20810},
      journal={arXiv preprint arXiv:2608.20810},
}

@article{xu2026lever,
  title = {Lever Can Move the Earth: Toward Adaptive Semantic Capacity Balance for Image-Text Retrieval},
  author = {Xu, Junhao and Liu, Zheng and Zhang, Mengqi and Dong, Guangyuan and Chen, Zhumin},
  journal = {IEEE Transactions on Multimedia},
  year = {2026},
  volume = {28},
  pages = {13369--13383},
  doi = {10.1109/TMM.2026.3723619},
  url = {https://doi.org/10.1109/TMM.2026.3723619}
}

@article{kang2026drilltrainingworldmodels,
      title={{DRILL}: Training World Models to Improve Policies, Not Predict Pixels}, 
      author={Zhaolu Kang and Tailong Luo and Lei Wei and Guangyuan Dong and Sihan Liu and Yingjie He and Siheng Wang and Shuaibo Li and Rongchao Zhang and Zhen Tian and Richeng Xuan and Zhichao Hu},
      year={2026},
      eprint={2610.30814},
      archivePrefix={arXiv},
      primaryClass={cs.CV},
      url={https://arxiv.org/abs/2610.30814},
      journal={arXiv preprint arXiv:2610.30814},
      note={Accepted at NeurIPS 2026},
}

@inproceedings{liu2025toward,
  title={Toward adaptive large language models structured pruning via hybrid-grained weight importance assessment},
  author={Liu, Jun and Kong, Zhenglun and Zhao, Pu and Yang, Changdi and Shen, Xuan and Tang, Hao and Yuan, Geng and Niu, Wei and Zhang, Wenbin and Lin, Xue and others},
  booktitle={Proceedings of the AAAI Conference on Artificial Intelligence},
  volume={39},
  pages={18879--18887},
  year={2025}
}

@article{tan2025perturbation,
  title={Perturbation-efficient zeroth-order optimization for hardware-friendly on-device training},
  author={Tan, Qitao and Chang, Sung-En and Xia, Rui and Ji, Huidong and Yang, Chence and Zhang, Ci and Liu, Jun and Zhan, Zheng and Zou, Zhou and Wang, Yanzhi and others},
  journal={arXiv preprint arXiv:2504.20314},
  year={2025}
}

@inproceedings{wang2025fairgnn,
  title={fairgnn-wod: Fair graph learning without complete demographics},
  author={Wang, Zichong and Liu, F and Pan, S and Liu, J and Saeed, F and Qiu, M and Zhang, W},
  booktitle={Proceedings of the 34th International Joint Conference on Artificial Intelligence},
  year={2025}
}

@inproceedings{yang2025fairsmoe,
  title={FairSMOE: Mitigating Multi-Attribute Fairness Problem with Sparse Mixture-of-Experts},
  author={Yang, Changdi and Zhan, Zheng and Zhang, Ci and Gong, Yifan and Liu, Jun and Shen, Xuan and Tang, Hao and Yuan, Geng and Zhao, Pu and Lin, Xue and Yanzhi, Wang},
  booktitle={Proceedings of the 34th International Joint Conference on Artificial Intelligence},
  year={2025}
}

@inproceedings{ji2025computation,
  title={A computation and energy efficient hardware architecture for ssl acceleration},
  author={Ji, Huidong and Li, Sheng and Cao, Yue and Ding, Chen and Xu, Jiawei and Tan, Qitao and Liu, Jun and Li, Ao and Tang, Xulong and Zheng, Lirong and others},
  booktitle={Proceedings of the 30th Asia and South Pacific Design Automation Conference},
  pages={23--29},
  year={2025}
}

@inproceedings{zhang2025towards,
  title={Towards memory-efficient and sustainable machine unlearning on edge using zeroth-order optimizer},
  author={Zhang, Ci and Yang, Chence and Tan, Qitao and Liu, Jun and Li, Ao and Wang, Yanzhi and Lu, Jin and Wang, Jinhui and Yuan, Geng},
  booktitle={Proceedings of the Great Lakes Symposium on VLSI 2025},
  pages={227--232},
  year={2025}
}

@inproceedings{liu2025rora,
  title={Rora: Efficient fine-tuning of llm with reliability optimization for rank adaptation},
  author={Liu, Jun and Kong, Zhenglun and Dong, Peiyan and Shen, Xuan and Zhao, Pu and Tang, Hao and Yuan, Geng and Niu, Wei and Zhang, Wenbin and Lin, Xue and others},
  booktitle={ICASSP 2025-2025 IEEE International Conference on Acoustics, Speech and Signal Processing (ICASSP)},
  pages={1--5},
  year={2025},
  organization={IEEE}
}

@article{chinta2025ai,
  title={AI-driven healthcare: Fairness in AI healthcare: A survey},
  author={Chinta, Sribala Vidyadhari and Wang, Zichong and Palikhe, Avash and Zhang, Xingyu and Kashif, Ayesha and Smith, Monique Antoinette and Liu, Jun and Zhang, Wenbin},
  journal={PLOS Digital Health},
  volume={4},
  number={5},
  pages={e0000864},
  year={2025},
  publisher={Public Library of Science San Francisco, CA USA}
}

@inproceedings{meng2024instructgie,
  title={Instructgie: Towards generalizable image editing},
  author={Meng, Zichong and Yang, Changdi and Liu, Jun and Tang, Hao and Zhao, Pu and Wang, Yanzhi},
  booktitle={European Conference on Computer Vision},
  pages={18--34},
  year={2024},
  organization={Springer}
}

@inproceedings{chinta2023optimization,
  title={Optimization and improvement of fake news detection using voting technique for societal benefit},
  author={Chinta, Sribala Vidyadhari and Fernandes, Karen and Cheng, Ningxi and Fernandez, Jordan and Yazdani, Shamim and Yin, Zhipeng and Wang, Zichong and Wang, Xuyu and Xu, Weifeng and Liu, Jun and others},
  booktitle={2023 IEEE International Conference on Data Mining Workshops (ICDMW)},
  pages={1565--1574},
  year={2023},
  organization={IEEE}
  }

@article{10.1145/3747842,
author = {Niu, Wei and Sun, Mengshu and Li, Zhengang and Chen, Jou-An and Guan, Jiexiong and Shen, Xipeng and Liu, Jun and Zhang, Mei and Wang, Yanzhi and Lin, Xue and Ren, Bin},
title = {Mobile-3DCNN: An Acceleration Framework for Ultra-Real-Time Execution of Large 3D CNNs on Mobile Devices},
year = {2025},
publisher = {Association for Computing Machinery},
address = {New York, NY, USA},
issn = {1544-3566},
url = {https://doi.org/10.1145/3747842},
doi = {10.1145/3747842},
journal = {ACM Trans. Archit. Code Optim.},
month = jul
}

@inproceedings{yuan2021work,
  title={Work in progress: Mobile or FPGA? A comprehensive evaluation on energy efficiency and a unified optimization framework},
  author={Yuan, Geng  and others},
  booktitle={IEEE Real-Time and Embedded Technology and Applications Symposium (RTAS)},
  pages={493--496},
  year={2021},
 
}

@article{yuan2022mobile,
  title={Mobile or FPGA? A Comprehensive Evaluation on Energy Efficiency and a Unified Optimization Framework},
  author={Yuan, Geng and Dong, Peiyan and Sun, Mengshu and others},
  journal={ACM Transactions on Embedded Computing Systems},
  volume={21},
  number={5},
  pages={1--22},
  year={2022},
}

@article{niu2025mobile,
  title={Mobile-3dcnn: An acceleration framework for ultra-real-time execution of large 3d cnns on mobile devices},
  author={Niu, Wei and Sun, Mengshu and Li, Zhengang and Chen, Jou-An and Guan, Jiexiong and Shen, Xipeng and Liu, Jun and Zhang, Mei and Wang, Yanzhi and Lin, Xue and others},
  journal={ACM Transactions on Architecture and Code Optimization},
  volume={22},
  number={3},
  pages={1--22},
  year={2025},
  publisher={ACM New York, NY}
}

@article{tan2025zeroqat,
  title={ZeroQAT: Your Quantization-aware Training but Efficient},
  author={Tan, Qitao and Song, Xiaoying and Lu, Jin and Li, Guoming and Liu, Jun and Hong, Lingzi and Ding, Caiwen and Li, Jundong and Zhai, Xiaoming and Huang, Shaoyi and others},
  journal={arXiv preprint arXiv:2509.00031},
  year={2025}
}

@inproceedings{li2025mutual,
  title={Mutual Effort for Efficiency: A Similarity-based Token Pruning for Vision Transformers in Self-Supervised Learning},
  author={Li, Sheng and Tan, Qitao and Dai, Yue and Kong, Zhenglun and Wang, Tianyu and Liu, Jun and Li, Ao and Liu, Ninghao and Ding, Yufei and Tang, Xulong and others},
  booktitle={The Thirteenth International Conference on Learning Representations},
  year={2025}
}

@inproceedings{liu2023scalable,
  title={A Scalable Real-time Semantic Segmentation Network for Autonomous Driving},
  author={Liu, Jun and Wu, Chao and Yuan, Geng and Niu, Wei and others},
  booktitle={Advanced Multimedia Computing for Smart Manufacturing and Engineering (AMC-SME)},
  pages={3--12},
  year={2023}
}

@article{liu2023interpretable,
  title={An Interpretable CNN for the Segmentation of the Left Ventricle in Cardiac MRI by Real-Time Visualization.},
  author={Liu, Jun and Yuan, Geng and Yang, Changdi and Song, Houbing and Luo, Liang},
  journal={CMES-Computer Modeling in Engineering \& Sciences},
  volume={135},
  number={2},
  year={2023}
}

@inproceedings{liu2021explainable,
  title={An Explainable Convolutional Neural Networks for Automatic Segmentation of the Left Ventricle in Cardiac MRI.},
  author={Liu, Jun and Deng, Feng and Yuan, Geng and others},
  booktitle={CECNet},
pages={306--314},
  year={2021}
}

@article{liu2022efficient,
  title={An Efficient CNN for Radiogenomic Classification of Low-Grade Gliomas on MRI in a Small Dataset},
  author={Liu, Jun and Deng, Feng and Yuan, Geng and Yang, Changdi and others},
  journal={Wireless Communications and Mobile Computing},
  volume={2022},
  number={1},
  year={2022}
}

@article{liu2024brain,
  title={Brain Tumor Classification on MRI in Light of Molecular Markers},
  author={Liu, Jun and Yuan, Geng and Zeng, Weihao and Tang, Hao and Zhang, Wenbin and others},
  journal={arXiv preprint arXiv:2409.19583},
  year={2024}
}

@article{wang2025graph,
  title={Graph Fairness via Authentic Counterfactuals: Tackling Structural and Causal Challenges},
  author={Wang, Zichong and Yin, Zhipeng and Zhang, Yuying and Yang, Liping and Zhang, Tingting and Pissinou, Niki and Cai, Yu and Hu, Shu and Li, Yun and Zhao, Liang and others},
  journal={ACM SIGKDD Explorations Newsletter},
  volume={26},
  number={2},
  pages={89--98},
  year={2025},
  publisher={ACM New York, NY, USA}
}

@incollection{wang2025amcr,
  title={AMCR: A Framework for Assessing and Mitigating Copyright Risks in Generative Models},
  author={Zhipeng Yin and Zichong Wang and Avash Palikhe and Zhen Liu and Jun Liu and Wenbin Zhang},
  booktitle={ECAI 2025},
  year={2025},
  publisher={IOS Press}
}

@article{li2025local,
  title={From Local Cues to Global Percepts: Emergent Gestalt Organization in Self-Supervised Vision Models},
  author={Li, Tianqin and Wen, Ziqi and Song, Leiran and Liu, Jun and Jing, Zhi and Lee, Tai Sing},
  journal={arXiv preprint arXiv:2506.00718},
  year={2025}
}

@inproceedings{dong2024physical,
  title={Physical-space Multi-body Mesh Detection Achieved by Local Alignment and Global Dense Learning},
  author={Dong, Haoye and Xiang, Tiange and Chittupalli, Sravan and Liu, Jun and Huang, Dong},
  booktitle={WACV},
  pages={1267--1276},
  year={2024}
}

@inproceedings{dong2024df,
  title={DF-VTON: Dense Flow Guided Virtual Try-On Network},
  author={Dong, Haoye and Liu, Jun and Huang, Dong},
  booktitle={ICASSP},
  pages={3175--3179},
  year={2024},
  organization={IEEE}
}

@inproceedings{liu-etal-2026-rcr,
    title = "{RCR}-Router:Role-Conditioned Context Routing for Resource-Efficient Multi-Agent {LLM} Systems",
    author = "Liu, Jun  and
      Zhao, Pu  and
      Kong, Zhenglun  and
      Zhang, Wenbin  and
      Yang, Fan  and
      Dong, Peiyan  and
      Yang, Changdi  and
      Fan, Wuzhe  and
      Li, Tianqin  and
      Tang, Hao  and
      Yuan, Geng  and
      Niu, Wei  and
      Nanjekye, Joannah  and
      Lin, Xue  and
      Huang, Dong  and
      Wang, Yanzhi",
    editor = "Lee, Hung-Yi  and
      Dev, Sunipa  and
      Li, Jessy  and
      Elliott, Desmond",
    booktitle = "Findings of the {A}ssociation for {C}omputational {L}inguistics: {EMNLP} 2026",
    month = oct,
    year = "2026",
    address = "Budapest, Hungary",
    publisher = "Association for Computational Linguistics",
    url = "https://aclanthology.org/2026.findings-emnlp.425/",
    doi = "10.18653/2026.findings-emnlp.425",
    pages = "8938--8963",
   }

@article{liu2026should,
  title={When Should a Robot Think? Resource-Aware Reasoning via Reinforcement Learning for Embodied Robotic Decision-Making},
  author={Liu, Jun and Zhao, Pu and Kong, Zhenglun and Shen, Xuan and Dong, Peiyan and Yang, Fan and Cui, Lin and Tang, Hao and Yuan, Geng and Niu, Wei and others},
  journal={arXiv preprint arXiv:2603.16673},
  year={2026}
}

@inproceedings{liu2026structured,
  title={Structured Agent Distillation for Large Language Model Agents},
  author={Liu, Jun and Kong, Zhenglun and Dong, Peiyan and Yang, Changdi and Li, Tianqin and Xie, Yanyue and Gong, Yifan and Shen, Xuan and Zhao, Pu and Tang, Hao and others},
  booktitle={Proc. of the 25th International Conference on Autonomous Agents and Multiagent Systems},
  pages={3676--3685},
  year={2026}
}

@article{yin2026assisting,
  title={Assisting patients with reliable information gathering: A guide to building efficient QA systems for medical use},
  author={Yin, Zhipeng and Wang, Zichong and Chen, Min and Stockwell, Ian and Ning, Xin and Liu, Jun and Zhang, Wenbin},
  journal={PLOS Digital Health},
  volume={5},
  number={9},
  pages={e0001634},
  year={2026},
  publisher={Public Library of Science, San Francisco, CA USA}
}

@InProceedings{Zhang_2026_CVPR,
    author    = {Zhang, Ci and Ding, Zhaojun and Yang, Chence and Liu, Jun and Zhai, Xiaoming and Huang, Shaoyi and Li, Beiwen and Ma, Xiaolong and Lu, Jin and Yuan, Geng},
    title     = {Roots Beneath the Cut: Uncovering the Risk of Concept Revival in Pruning-Based Unlearning for Diffusion Models},
    booktitle = {Proceedings of the IEEE/CVF Conference on Computer Vision and Pattern Recognition (CVPR)},
    month     = {June},
    year      = {2026},
    pages     = {35872-35881}
}

@inproceedings{wang2026learning,
  title={Learning counterfactual fairness from authentic generation},
  author={Wang, Zichong and Yin, Zhipeng and Chen, Zhong and Yang, Jack and Liu, Jun and Zhang, Wenbin},
  booktitle={Proceedings of the Thirty-Fifth International Joint Conference on Artificial Intelligence},
  pages={798--806},
  year={2026},
  organization={International Joint Conferences on Artificial Intelligence Organization}
}
}

\clearpage
\setcounter{page}{1}
\maketitlesupplementary

\section{Timing Measurement Details}
\label{supp:timing}

This supplement documents the measurement setup behind
Table~\ref{tab:timing-breakdown} and the single-forward mechanism that explains
why per-forward savings do not carry the whole query. All figures are directly
measured; none are composed from component times.

\paragraph{Environment.}
NVIDIA RTX A6000, PyTorch 2.7.0 with CUDA 12.8 and cuDNN 90701, model
nvidia/Cosmos-Policy-RoboCasa-Predict2-2B with weights in bf16, latent shape
$[1,16,11,28,28]$ (11 latent frames $\times$ 196 tokens $=$ 2156), batch 1, and
5 denoising steps per action query (confirmed as exactly 5 DiT forwards per
call). The only tracked code modification is a logging patch to
\texttt{run\_robocasa\_eval.py} (commit \texttt{18a2acc}).

\paragraph{End-to-end protocol (Table~\ref{tab:timing-breakdown}b).}
For each pruning level and each of the two paths (deployed, and
decode-excluded), we time the complete action-inference query, dense against
pruned, under the deployed eager backend. We discard warmup and then record
$N=30$ timed queries per cell, reporting the median with the $10$th and $90$th
percentiles; the directory \texttt{out\_e2e\_prune/raw.jsonl} holds all 420
timed calls. Paired per-state speedups agree with the reported median ratios to
within $0.001$. The decode-excluded dense baseline is directly measured at
$812.7$ ms (it is not the decode-subtracted estimate used before this run). The
whole-frame prune is applied inside the real query path; because whole-frame
pruning is a fixed structural mask, there is no per-token scoring step, and the
frame slicing and scatter-back run inside the timed region, so the measured
speedups carry full real overhead for this pruning type.

\paragraph{Why the whole query was measured eager.}
The complete query contains host-side control flow, host-device copies and array
post-processing that cannot be CUDA Graph captured without restructuring the
code, so no graph-timed stage is mixed into the end-to-end numbers. Eager is in
any case the deployed evaluation backend.

\paragraph{Single-forward mechanism (not an end-to-end result).}
To locate the ceiling of the pruned stage, we separately time the DiT forward in
isolation. Under CUDA Graph the single forward speeds up by at most $1.73\times$
(dense $102.75$ ms to all-future-frames-plus-value $59.42$ ms), with output
bit-identical to eager. This is a single-forward, CUDA-Graph measurement and is
not directly comparable to the eager end-to-end speedups in
Table~\ref{tab:timing-breakdown}b; we report it only to show that even a forward
driven nearly to its floor cannot carry the query, because the five DiT forwards
are under half of the end-to-end latency (Table~\ref{tab:timing-breakdown}c)
while VAE encode and future-image decode are not reducible by token pruning.

\paragraph{FLOPs.}
Per-forward FLOPs in Table~\ref{tab:timing-breakdown}a are measured with
FlopCounterMode (matmul $+$ SDPA).

\paragraph{Reproduction.}
The summary is regenerated from the raw timings by
\texttt{E2E\_MODE=summarize python3 probes/ARC\_gate\_m1/gm1\_e2e\_prune.py};
\texttt{out\_e2e\_prune/} also contains \texttt{E2E\_PRUNE\_REPORT.md},
\texttt{summary.json} and \texttt{meta.json}.


\section{Statistical Definitions}
\label{supp:stat}

\paragraph{Decision threshold.}
The pre-registered interaction threshold is
\begin{equation}
\theta_B = \max\!\big(2\sigma_{\text{seed}},\; 0.3\, D(2,F)\big),
\label{eq:theta}
\end{equation}
the larger of a seed-noise floor and a fraction of the temporal main effect, so
that a declared interaction must exceed both the noise and a meaningful share of
the main effect.

\paragraph{Oracle headroom reliability.}
With $R$ replicate rollouts per state, the reliability of the mean candidate
estimate follows the Spearman--Brown form
\begin{equation}
\mathrm{Rel}(R) = \frac{R\,\rho_{\text{ICC}}}{1 + (R-1)\,\rho_{\text{ICC}}}.
\label{eq:reliability}
\end{equation}
With the measured intraclass correlation $\rho_{\text{ICC}} = 0.20$ and $R = 2$,
$\mathrm{Rel} = 0.33$, so a single continuation ranks candidates with only about
$20\%$ reliability and the cross-fitted oracle estimate is itself noisy.

\section{The Premise-First Audit Procedure}
\label{supp:methodology}

\begin{figure*}[t]
\centering
\includegraphics[width=0.92\linewidth]{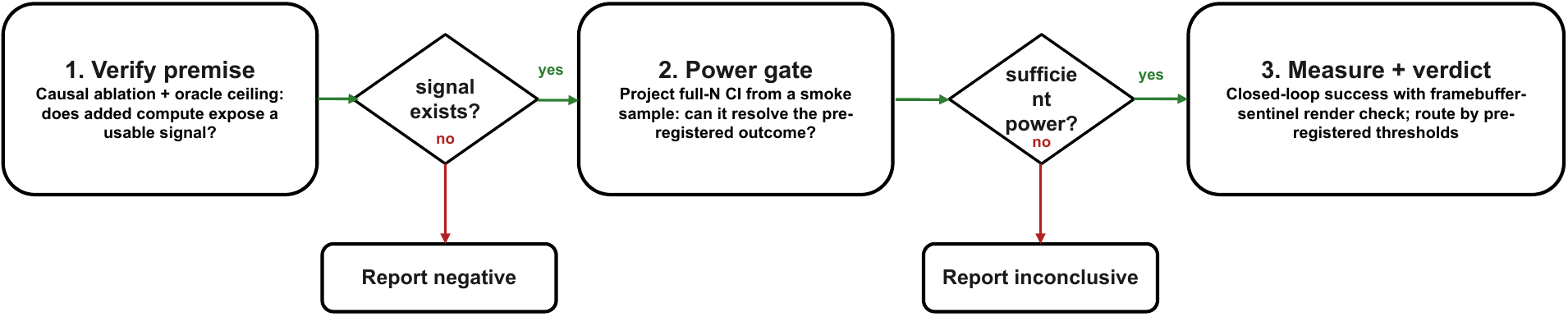}
\caption{The premise-first methodology as a chain of gates. Each candidate
test-time-scaling gain passes three checks in order before it is credited:
verify that the decision signal exists (causal ablation and oracle ceiling); a
statistical power gate that halts a run which cannot resolve the pre-registered
outcome; and closed-loop measurement under framebuffer-sentinel render-integrity
checking. A ``no'' at the first gate reports a negative result and at the second
an inconclusive one, separating a genuine negative from an underpowered or
invalid measurement.}
\label{fig:methodology}
\end{figure*}

Figure~\ref{fig:methodology} shows the audit as a single procedure: three gates
applied in order before any test-time-scaling gain is credited. Each gate
corresponds to one of the three ways an apparent gain can fail, and to one of
the paper's three audits.

The first gate verifies the premise that the added computation exposes a usable
decision signal, using the causal ablation and oracle-ceiling construction of
Section~\ref{sec:reliability}; when no signal exists the result is reported as negative rather than
pursued further, as in the planning audit of Section~\ref{sec:planning-audit}. The second gate is
statistical: from a smoke sample it projects the confidence-interval width at the
full sample size and halts the run when the pre-registered outcome cannot be
resolved, so compute is not spent on an experiment that cannot yield a verdict,
and the outcome is reported as inconclusive rather than as evidence of absence,
as in the allocation audit of Section~\ref{sec:allocation}. The third gate measures closed-loop
success under the framebuffer-sentinel render check and routes the verdict by
pre-registered thresholds, the render-integrity guard whose necessity the
reliability audit of Section~\ref{sec:reliability} establishes.

The value of the procedure is in its two ``no'' branches: they separate a
genuine negative from an underpowered or invalid measurement, a distinction that
success rate alone cannot make. The gates are model-agnostic and apply to any
audit of test-time compute in a diffusion-based control policy; only the render
fix itself is specific to the stack we study.


\section{End-to-End Latency Under Token Pruning}
\label{supp:e2e-detail}

Table~\ref{tab:e2e-detail} reports the full measurement behind
Table~\ref{tab:timing-breakdown}(b): both query paths at all six pruning levels,
with the 10th and 90th percentiles over the $N=30$ timed queries per cell. Three
observations follow, all from directly measured values.

First, the end-to-end speedup grows monotonically but sublinearly with the
nominal FLOPs reduction. Cutting $47.9\%$ of per-forward FLOPs (all future frames
plus value) buys only $15.2\%$ of wall-clock on the deployed path and $25.3\%$ on
the decode-excluded path; the remaining time is dominated by the autoencoder
encode and future-image decode, which token pruning does not touch.

Second, the decode-excluded path scales better at every level, reaching
$1.253\times$ against $1.152\times$ on the deployed path at the most aggressive
prune. Removing the fixed future-image decode shrinks the denominator and
enlarges the share of the query that pruning can act on, yet even this
most-favorable configuration stops well short of the $1.5\times$ target.

Third, the percentile spread is tight: the 10th and 90th percentiles lie within
about $1\%$ of the median at every cell, so the medians are stable and the
sub-$1.5\times$ conclusion is not an artifact of run-to-run variance. No pruning
level on either path reaches $1.5\times$.

\begin{table*}[t]
\centering
\footnotesize
\setlength{\tabcolsep}{5pt}
\caption{Full end-to-end query latency under whole-frame token pruning, both
paths, all levels (Cosmos Policy, RoboCasa, batch 1, 5 denoising steps; NVIDIA
RTX A6000, bf16, eager). Every value is directly measured over $N=30$ timed
queries per cell; median with the 10th and 90th percentiles. The deployed path
is the query as the eval calls it (future-image decode and value extraction
included); the decode-excluded path removes future-image decode on both sides.
Speedup is dense median divided by pruned median. No pruning level on either
path reaches the pre-registered $1.5\times$ target. This expands the medians and
speedups of main-body Table~\ref{tab:timing-breakdown}(b).}
\label{tab:e2e-detail}
\begin{tabular}{lc ccc ccc}
\toprule
& & \multicolumn{3}{c}{Deployed path} & \multicolumn{3}{c}{Decode-excluded path} \\
\cmidrule(lr){3-5}\cmidrule(lr){6-8}
Pruning level & Tokens & Median (ms) & p10--p90 (ms) & $\times$ & Median (ms) & p10--p90 (ms) & $\times$ \\
\midrule
Dense                       & 2156 & 1254.9 & 1248.9--1257.1 & 1.000 & 812.7 & 808.6--814.5 & 1.000 \\
Drop 1 future image         & 1960 & 1232.1 & 1226.7--1233.8 & 1.019 & 791.2 & 786.7--792.5 & 1.027 \\
Drop 2 future images        & 1764 & 1171.0 & 1168.7--1172.6 & 1.072 & 728.9 & 727.0--730.6 & 1.115 \\
All future images           & 1568 & 1140.5 & 1136.6--1142.8 & 1.100 & 699.8 & 697.0--703.2 & 1.161 \\
All future frames           & 1372 & 1096.5 & 1093.9--1099.3 & 1.144 & 654.0 & 652.0--656.5 & 1.243 \\
All future frames $+$ value & 1176 & 1088.9 & 1086.4--1092.2 & \textbf{1.152} & 648.6 & 647.0--651.9 & \textbf{1.253} \\
\bottomrule
\end{tabular}
\end{table*}

\section{Candidate Diversity and Outcome Separability}
\label{supp:diversity}

These diagnostics address whether the limited selection headroom in Section~4 is
an artifact of near-identical candidates or of replicate noise. They are
descriptive and read only saved files (candidate actions, per-candidate closed
outcomes with two repeats, and value scores); they do not run inference, and they
do not claim that greater diversity would improve success.

\paragraph{Candidates are diverse, not degenerate.}
For each state we z-score the seven action dimensions across the dataset, flatten
each candidate's executed 16-step chunk to a 112-vector, and take the mean
pairwise Euclidean distance among the eight candidates. No state has identical
candidates (minimum dispersion $0.18$). States without outcome variation show
\emph{greater} absolute dispersion than states with it (pooled median $0.78$ vs
$0.48$; Mann--Whitney $p=5\times10^{-5}$; Spearman $\rho=-0.25$ against outcome
degree), a pattern driven mainly by TurnOnStove. Scale-free coefficient-of-
variation metrics show no significant group difference ($p=0.07$--$0.10$), so the
difference is largely one of absolute action magnitude. Value-score dispersion is
uniformly tiny (median standard deviation $\approx 3\times10^{-4}$) and unrelated
to outcome variation in the pooled analysis ($p=0.62$), consistent with a value
selector that carries little usable ranking signal; TurnOnStove, where the value
selector is worst, is the only task with larger value dispersion on varying
states.

\paragraph{Decomposing outcome variation.}
The $48$ of $263$ states with any variation (across eight candidates and two
repeats) split into those with a stable between-candidate difference and those
without one. We call a state stably separable when some candidate succeeds in
both repeats while another fails in both.

\begin{table*}[h]
\centering
\small
\begin{tabular}{@{}lccccc@{}}
\toprule
Task & States & Varying & Stably sep. & No stable sep. & No variation \\
\midrule
OpenSingleDoor  & 37  & 0  & 0  & 0  & 37 \\
PnPCounterToCab & 78  & 10 & 6  & 4  & 68 \\
TurnOnStove     & 76  & 33 & 14 & 19 & 43 \\
TurnOffStove    & 72  & 5  & 0  & 5  & 67 \\
\midrule
\textbf{Pooled} & 263 & 48 & \textbf{20} & \textbf{28} & 215 \\
\bottomrule
\end{tabular}
\caption{Of $48$ varying states, only $20$ ($7.6\%$) have a stable
between-candidate outcome difference; the other $28$ exhibit variation without
satisfying the stable-separation criterion. All $28$ are graded ($0.5$ vs $0$ or
$0.5$ vs $1$); none is a full-repeat flip. Repeat disagreement concentrates in
TurnOnStove (per-candidate repeat agreement $0.83$ vs $0.97$--$1.00$ elsewhere).}
\label{tab:noise-decomp}
\end{table*}

\paragraph{Even the stable states are near chance.}
A within-state null that independently shuffles the candidate labels in each
repeat (preserving each repeat's success count but breaking candidate identity
across repeats, $2000$ draws) expects $\approx 14.8$ stably separable states by
chance, against the observed $20$; the excess is about $5$ states, mostly
PnPCounterToCab. The cross-fitted hindsight oracle (same estimator as the main
text) gains $+1.40$\,pp pooled, decomposing as $+1.85$\,pp from the $20$ stably
separable states and $-0.44$\,pp from the $28$ states without stable separation
(on which the cross-fitted oracle never beats random). Under this specific
label-shuffle null the pooled cross-fitted oracle has a near-zero mean gain
($+0.01$\,pp), providing no indication of upward inflation under the null; with
only two repeats per candidate, the amount of reproducible selection headroom
remains uncertain. This contribution split is exploratory: the grouping is itself
defined from the two repeats and is subject to sampling randomness.

\paragraph{Limitations.}
Actions cover only the executed first $16$ of the $32$ generated steps. With two
repeats per candidate we cannot adjudicate per state whether a stable separation
is real or coincidental; the graded ($0.5$-valued) states are indistinguishable
from noise at this sample size. States within an episode are not independent and
the reported $p$-values are not corrected for multiple comparisons, so they are
exploratory rather than confirmatory. The diagnostics cover four tasks and three
episodes and do not speak to other candidate counts, sampling temperatures, or
continuation policies. Full numbers and reproducible scripts are in
\texttt{probes/P5\_oracle/} (\texttt{p5\_diversity.py}, \texttt{p5\_noise\_decomp.py}).

\end{document}